\documentclass[runningheads]{llncs}
\usepackage{times}
\usepackage{latexsym}
\usepackage[subtle]{savetrees}
\usepackage[T1]{fontenc}

\usepackage[utf8]{inputenc}

\usepackage{microtype}

\usepackage{inconsolata}

\usepackage{graphicx}
\usepackage{amsfonts}
\usepackage{amsmath}
\usepackage{bbm}
\usepackage{booktabs}
\usepackage{array}

\usepackage{wrapfig}
\usepackage{inconsolata}
\usepackage{xspace}
\usepackage{graphicx}
\usepackage{booktabs} 
\usepackage{marvosym}  
\usepackage{graphicx}
\usepackage{caption}
\usepackage{stfloats} 
\usepackage{cuted}
\usepackage{multirow}
\usepackage{tabularx}
\usepackage{subcaption}
\usepackage{flushend}

\usepackage{algorithm}
\usepackage{algpseudocode}

\usepackage[flushleft]{threeparttable}
\usepackage{scalerel,xparse}
\usepackage{hyperref}
\usepackage{color}
\usepackage{wrapfig}

\usepackage{tocloft}
\usepackage{etoc}

\usepackage{adjustbox}
\usepackage{enumitem}

\usepackage[table]{xcolor}
\usepackage{lipsum}
\usepackage{titletoc}
\usepackage{tcolorbox}
\usepackage{arydshln}
\usepackage{comment}

\tcbuselibrary{breakable}

\newcommand{\ourwork}{\textsc{AgentActionBench}}
\newcommand{\ourworksmall}{\textsc{AgentActionBench-Human}}

\newif\ifreview
\reviewtrue       

\newcommand{\tcbtab}{\hspace*{2ex}}
\newcommand{\hyphentt}[1]{\texttt{\hyphenchar\font=\defaulthyphenchar #1}}

\algrenewcommand\algorithmicrequire{\textbf{Input:}}
\algrenewcommand\algorithmicensure{\textbf{Output:}}

\expandafter\def\expandafter\normalsize\expandafter{%
    \normalsize%
    \setlength\abovedisplayskip{0pt}%
    \setlength\belowdisplayskip{2pt}%
}
\begin{document}
\title{Overview of the NLPCC 2026 Shared Task 11: Agent-Based Experiment Reproduction from Scientific Papers}
\titlerunning{NLPCC 2026 Shared Task 11}
%
%
%
\author{
Hanhua Hong\inst{1} \and
Yizhi Li\inst{2} \and
Luu Gia Huy\inst{3} \and
Jian Yang\inst{4} \and
Ming Zhou\inst{5} \and
Chenghua Lin\inst{1} 
}
\authorrunning{H. Hong et al.}
%
 \institute{
\textsuperscript{1}The University of Manchester, 
 \textsuperscript{2}IQuest Research
 \\
 \textsuperscript{3}Vietnam National University
 \textsuperscript{4}Beihang University, 
 \textsuperscript{5}Langboat
 }

\maketitle              
\begin{abstract}
Reproducibility is essential to scientific progress, yet the growing volume and complexity of scientific publications make exhaustive manual verification increasingly impractical. Although recent advances in large language model (LLM) agents enable automated experiment reproduction, existing evaluations largely focus on final repositories and are typically limited to machine learning (ML). We introduce \ourwork, a process-oriented benchmark for evaluating agent-based experiment reproduction across ML and AI4Science domains. Our framework uses an MCP-based Action Recorder to capture agents' behaviour throughout the reproduction process and evaluates the resulting traces with paper-specific rubrics. \ourwork~contains 150 papers, including 120 ML papers and 30 AI4Science papers. A human-annotated subset covering 10\% of the benchmark provides validation data, while model-assisted augmentation expands the full benchmark to more than 10,000 rubric items. Experimental results show that current systems remain limited, with execution as the primary bottleneck. Meanwhile, the strong Pearson and Spearman correlations between model-generated and human-annotated rubrics validate the reliability of our scalable rubric-generation approach.
\end{abstract}
\section{Introduction}

Reproducibility is essential to trustworthy scientific progress, yet it remains difficult to achieve in practice. A \textit{Nature} survey found that more than 70\% of researchers had failed to reproduce another scientist's experiments, while recent studies indicate that only around 20\% of papers at top computer science conferences release code sufficient for reproduction~\cite{baker20161,doi:10.1142/S0218194022500358,magnusson-etal-2023-reproducibility}. As scientific publications grow in both volume and complexity, manual verification at scale is becoming increasingly impractical.

Recent advances in large language models (LLMs) have substantially improved long-context understanding, code generation, reasoning, and tool use. Consequently, LLM-based agents are increasingly applied to complex scientific workflows, from machine learning experimentation to AI for Science (AI4Science) tasks in domains such as biology and chemistry~\cite{abramson_accurate_2024,bran2023chemcrowaugmentinglargelanguagemodels,hong2025onesizefitsallinversionlearninghighly,lyu2026evoscientistmultiagentevolvingai,yuan-etal-2025-dolphin,zhao2025autoreproduceautomaticaiexperiment}. These capabilities make them promising candidates for automated experiment reproduction, which requires paper comprehension, implementation planning, code generation, experiment execution, and result verification.

\begin{figure}[htb]
    \centering
    \includegraphics[width=0.9\linewidth]{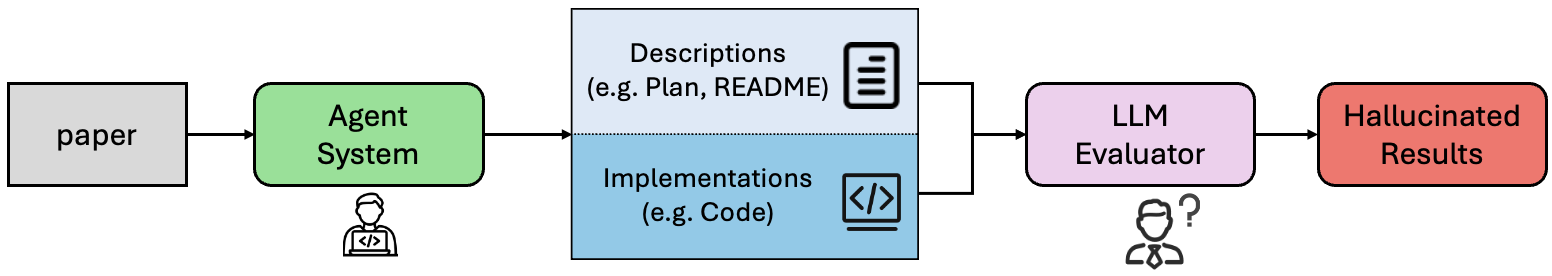}
    \caption{Illustration of output-oriented evaluation in prior benchmarks. When descriptions and implementations are evaluated together, LLM-based evaluators may incorrectly infer implementations that were never produced or executed.}
    \label{fig:previous}
    \vspace{-2em}
\end{figure}

Recent benchmarks, including Paper2Code~\cite{seo2025paper2codeautomatingcodegeneration} and PaperBench~\cite{starace2025paperbenchevaluatingaisability}, evaluate whether AI systems can reproduce research papers. However, most of them focus on machine learning and assess final artefacts rather than the reproduction process. Such output-oriented evaluation can be unreliable since an agent may generate plausible files or descriptions while overlooking important paper content, failing to execute its code, or fabricating execution records. As illustrated in Fig.~\ref{fig:previous}, this ambiguity makes it difficult to determine whether a reproduction is grounded in the source paper and supported by concrete actions.

\begin{figure}[htb]
    \vspace{-4em}
    \centering
    \includegraphics[width=\linewidth]{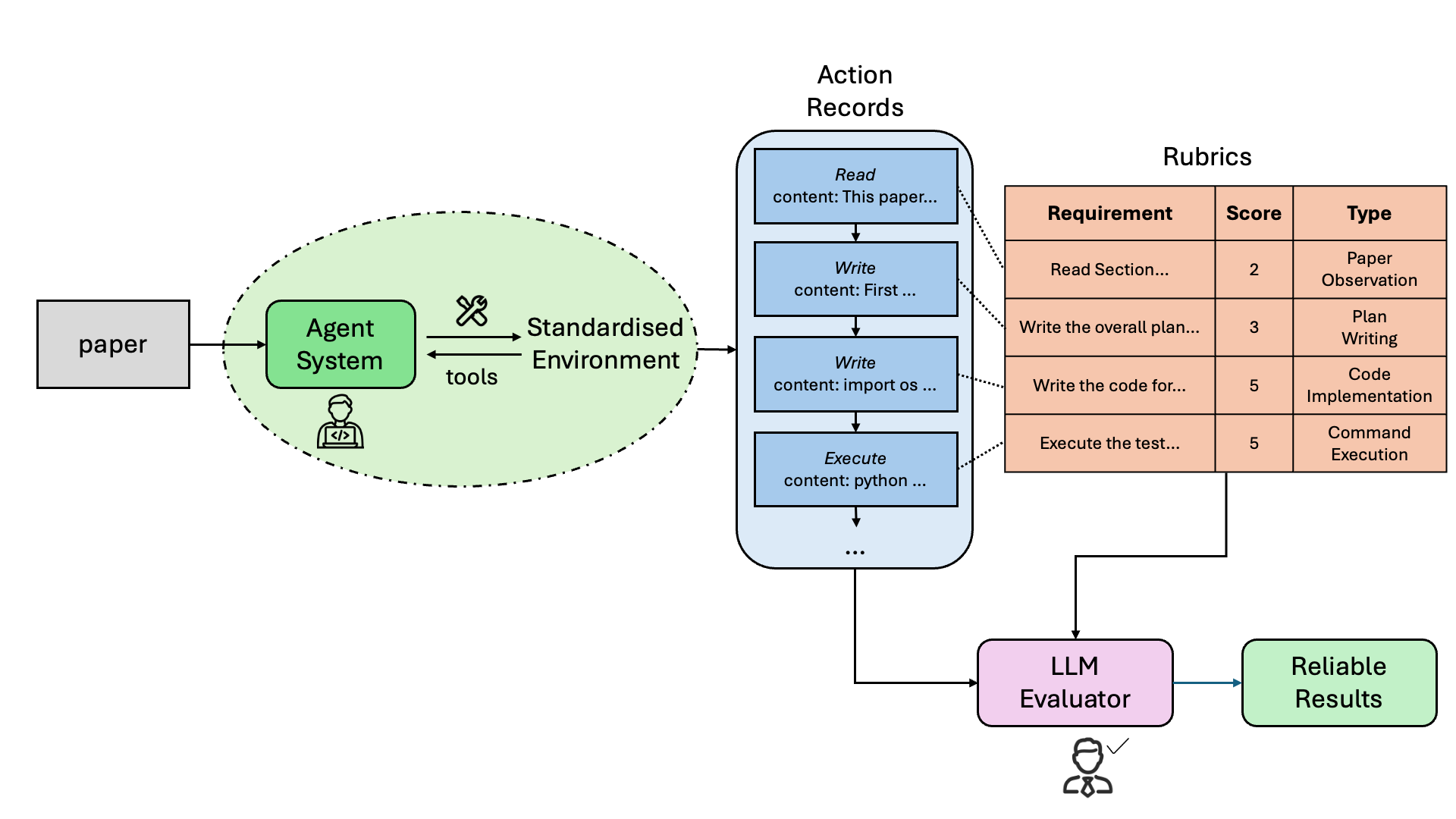}
    \vspace{-1.5em}
    \caption{Overview of our framework. A standardised execution environment records agent actions, which are matched against paper-specific rubrics for auditable, process-oriented evaluation.}
    \label{fig:ours}
    \vspace{-2em}
\end{figure}

To address these limitations, we introduce a process-oriented benchmark for agent-based experiment reproduction. Rather than evaluating only the final repository, our framework decomposes reproduction into fine-grained stages and verifies whether agents perform the required operations. An MCP-based Action Recorder captures reading, writing, and command-execution actions within a standardised environment. The resulting action logs provide auditable evidence for rubric-based evaluation, allowing each stage of the reproduction workflow to be assessed directly. Fig.~\ref{fig:ours} presents an overview of the framework.

We further extend evaluation beyond conventional machine learning papers to AI4Science research in astronomy, biology, chemistry, environmental science, materials science, and medicine. We first construct \ourworksmall, a manually annotated subset containing 12 ML and 3 AI4Science papers. To reduce the cost of expert annotation, we develop an agent-assisted rubric-generation pipeline, validate it against the human annotations in \ourworksmall, and scale the benchmark to the full \ourwork, comprising 120 ML and 30 AI4Science papers with more than 10,000 rubric items.

In summary, our contributions are threefold:

\setlist{nolistsep}
\begin{itemize}

\item We introduce \ourwork, a process-oriented, multi-domain benchmark that evaluates the complete experiment-reproduction workflow across 150 ML and AI4Science papers.

\item We develop an MCP-based Action Recorder and a rubric-grounded evaluation protocol that use auditable action traces to assess paper understanding, implementation, execution, and result verification.

\item We evaluate systems submitted to the NLPCC 2026 shared task, revealing key limitations of current agents in experiment reproduction and validating the reliability of our scalable rubric-generation pipeline.

\end{itemize}

\section{Related Work}
\subsection{Agent-Based Experiment Reproduction}
\label{sec:reproduction}

Autonomous scientific agents are increasingly studied for their ability to reproduce published experiments, an important capability for improving scientific efficiency and reproducibility. Existing systems address different stages of this process. The AI Scientist~\cite{DBLP:journals/corr/abs-2408-06292} supports an end-to-end workflow from idea generation to paper writing; DOLPHIN~\cite{yuan-etal-2025-dolphin} adopts a closed-loop process of ideation, implementation, and feedback; PaperCoder~\cite{seo2025paper2codeautomatingcodegeneration} converts research papers into executable repositories through a three-stage pipeline; and HiRAS~\cite{hong2026hirashierarchicalmultiagentframework} coordinates specialised sub-agents across fine-grained reproduction stages. A growing body of benchmarks evaluates these capabilities, including end-to-end ML reproduction~\cite{autoexperiment,,siegel2024corebench,starace2025paperbenchevaluatingaisability}, NLP-focused implementation from algorithmic descriptions~\cite{xiang2025scireplicate,yan-etal-2025-lmr}, and general ML-engineering tasks based on large-scale experiments and public competitions~\cite{chan2024mlebench,huang2024mlagentbench} Their results consistently show that current agents remain far behind human experts, particularly when tasks require robust implementation and execution. However, most existing evaluations focus on coarse outcome-level metrics, provide limited visibility into the intermediate reproduction process, and primarily target a single domain. In contrast, our benchmark evaluates the complete workflow from paper understanding to result verification through auditable action traces, enabling fine-grained assessment across both ML and AI4Science papers.

\subsection{Rubric-Based Evaluation}
\label{sec:rubric}
Rubrics provide structured supervision by decomposing complex tasks into fine-grained evaluation criteria, addressing the limitations of coarse, outcome-based metrics~\cite{gunjal2026rubrics}. Benchmarks such as PaperBench~\cite{starace2025paperbenchevaluatingaisability} and HealthBench~\cite{arora2025healthbenchevaluatinglargelanguage} use human-authored rubrics for stepwise evaluation, but their reliance on domain experts makes them costly to construct and difficult to scale. This limitation has motivated model-generated rubrics~\cite{rubrichub2026,liu2026openrubricsscalablesyntheticrubric,zhang2026rubricbenchaligningmodelgeneratedrubrics}. For example, RubricHub~\cite{rubrichub2026} combines principle-guided synthesis, multi-model aggregation, and difficulty evolution to produce discriminative criteria, while prior work shows that question-specific rubrics improve the accuracy and consistency of logical evaluation~\cite{rubricisallyouneed2025}. Building on these studies, we generate rubrics specifically for research-paper reproduction. Unlike methods designed for general reward modelling or open-ended question answering, our approach targets the multi-stage, repository-grounded nature of experiment reproduction.

\subsection{LLM-Based Data Augmentation}
\label{sec:augmentation}
Large language models have increasingly been used to generate training and benchmark data themselves, driven by the high cost of manual annotation and the need to scale supervision alongside larger models~\cite{zhou2025textda}. \cite{zhou2025textda} survey this shift, organizing LLM-based augmentation methods into simple, prompt-based, and retrieval-based categories. One prominent line of work focuses on scaling instruction diversity to align models with human intent, from bootstrapping new instruction-response pairs from a small seed set~\cite{wang2023selfinstruct} to progressively rewriting instructions into more complex variants~\cite{xu2023wizardlm}. \cite{gao2023gllava} apply the same idea to a specific domain, prompting an LLM to expand existing geometry problem sets (GeoQA+, Geometry3K) into the larger Geo170K dataset. Together, these works show that LLM-driven generation can substitute for costly manual curation when scaling supervision or data diversity. We apply this same principle to automatically synthesise fine-grained, paper-specific rubrics for experiment reproduction, using an agent framework in place of the single-pass prompting adopted in prior augmentation work.
\begin{table}[t]
    \centering
    \caption{Tools provided by the Action Recorder. Optional arguments are enclosed in brackets.}
    \begin{adjustbox}{max width=\textwidth}
    \begin{tabular}{ccc}
    \toprule
    \textbf{Tool Name} & \textbf{Arguments} & \textbf{Results} \\
    \midrule
    \texttt{Read} & path: str$[$, start\_offset: int, end\_offset: int$]$ & content: str \\
    \texttt{Write} & path: str, content: str & success: bool \\
    \texttt{Execute} & cmd: str & success: bool, stdout: str, stderr: str \\
    \bottomrule
    \end{tabular}
    \end{adjustbox}
    \label{tab:tools}
    \vspace{-2em}
\end{table}
\section{Methodology}

\subsection{Action Recorder}
\label{sec:action_recorder}

To faithfully capture agent behaviour during experiment reproduction, we implement an Action Recorder based on the Model Context Protocol (MCP). Rather than granting agents direct access to the underlying operating system, the recorder mediates all interactions with the execution environment through a small set of MCP tools. Each operation is therefore captured automatically as a structured action log, providing a complete and reproducible trace of the reproduction process.

As summarised in Table~\ref{tab:tools}, the Action Recorder provides three core tools: \texttt{Read}, \texttt{Write}, and \texttt{Execute}. The \texttt{Read} tool retrieves file contents and optionally supports line-range selection for efficient inspection of large files. The \texttt{Write} tool creates or modifies source code, configuration files, and other artefacts required for reproduction. The \texttt{Execute} tool runs shell commands and returns the execution status, standard output, and standard error, enabling agents to conduct experiments and inspect their results. For safety and fairness, all tool interactions are restricted to a designated workspace.

The Action Recorder treats each tool invocation as an atomic, standardised action. For each invocation, it records the input arguments, returned results, and timestamp in a chronological JSON log, as illustrated in Appendix~\ref{app:log_example}. The log is automatically exported when the agent terminates. Because all interactions are mediated through the MCP interface, the recorder captures agent behaviour without modifying the agent and remains compatible with any MCP-enabled system. These structured logs provide the primary evidence for rubric-based evaluation, enabling fine-grained assessment of the reproduction process rather than relying solely on the final repository.

\subsection{Benchmark Construction}

We collect 150 papers to construct the full \ourwork~benchmark. Of these, 120 are drawn from top-tier machine learning conferences, including ICML, ICLR, ACL, and NeurIPS, while the remaining 30 are AI4Science papers published in leading scientific journals, such as \textit{Nature}, \textit{Bioinformatics}, and \textit{npj Computational Materials}.

We adopt a rubric-based evaluation protocol in which each paper is paired with a paper-specific checklist for assessing reproduction quality. Formally, each rubric item is represented as $r_i=(c_i,s_i,t_i)$, where $c_i$ denotes the evaluation criterion, $s_i$ its importance score, and $t_i$ the corresponding stage of the reproduction process. We define five rubric types: \textit{Paper Observation}, \textit{Plan Writing}, \textit{Code Implementation}, \textit{Command Execution}, and \textit{Result Matching}. Together, these categories cover the full reproduction workflow and enable a more comprehensive assessment than approaches that consider only the consistency between a generated repository and the source paper.

We construct the human-annotated subset, \ourworksmall, by randomly sampling 12 ML papers and 3 AI4Science papers, representing 10\% of the full benchmark. Three research students with extensive experience in publishing at and reviewing for top-tier conferences annotate the rubrics. After the initial annotation, each rubric set is independently reviewed by the other two annotators in rotation and revised as needed. This process continues until all annotators reach consensus. The resulting subset contains an average of 67.75 rubric items per paper and more than 1,000 manually annotated items in total.

Building on recent advances in LLM-based data augmentation, we scale the benchmark by automatically generating rubrics for all papers. We first convert the papers to Markdown using \hyphentt{MinerU}~\cite{wang2026mineru2}, providing LLMs with cleaner and more structured inputs. We then apply the rubric-generation framework introduced in prior work~\cite{hong2026llmswritereliablerubrics} to produce paper-specific rubrics, expanding \ourworksmall~into the full \ourwork~benchmark. The resulting benchmark contains more than 10,000 rubric items, with an average of 69.92 items per paper. Finally, we assess the alignment between model-generated and human-annotated rubrics on \ourworksmall~using correlation analysis. Detailed results are presented in \S\ref{sec:benchmark_analysis}.

\begin{table}[t]
    \centering
    \caption{Mapping between rubric types and the Action Recorder tools whose logs provide the corresponding evaluation evidence.}
    \begin{adjustbox}{max width=0.95\textwidth}
    \begin{tabular}{cc}
    \toprule
    \textbf{Rubric Type} & \textbf{Relevant Tools} \\
    \midrule
    \textit{Paper Observation} & \texttt{Read} \\
    \textit{Plan Writing} & \texttt{Write} \\
    \textit{Code Implementation} & \texttt{Write} \\
    \textit{Command Execution} & \texttt{Execute} \\
    \textit{Result Matching} & \texttt{Read, Execute} \\
    \bottomrule
    \end{tabular}
    \end{adjustbox}
    \label{tab:rubric_type}
    \vspace{-2em}
\end{table}

\subsection{Evaluation}

Our evaluation framework primarily relies on the action log $L$ exported by the Action Recorder described in \S\ref{sec:action_recorder}. For each paper $\mathcal{P}$, the evaluator processes every rubric item $r_i=(c_i,s_i,t_i)$. To reduce context length and exclude irrelevant evidence, the evaluator model $\mathcal{M}$ first extracts a subset of logs $L_i' \subseteq L$ associated with the tools relevant to rubric type $t_i$. Table~\ref{tab:rubric_type} specifies the mapping between each rubric type and its corresponding tools.

Given the extracted logs $L_i'$, rubric criterion $c_i$, and source paper $\mathcal{P}$, the evaluator assigns a binary judgment $J_i \in \{\mathrm{Pass},\mathrm{Fail}\}$ as $J_i=\mathcal{M}(L_i',c_i,\mathcal{P})$. The overall reproduction score is then computed as the importance-weighted proportion of passed rubric items:
\begin{equation}
    \mathrm{score}
    =
    \frac{
        \sum_i \left[J_i=\mathrm{Pass}\right] s_i
    }{
        \sum_i s_i
    },
\end{equation}
where $[\cdot]$ equals $1$ when the enclosed condition is satisfied and $0$ otherwise.

\begin{table}[t]
    \centering
    \caption{Leaderboard results for the shared task. Our reproduction baseline is underscored.}
    \begin{adjustbox}{max width=\textwidth}
    \begin{tabular}{cccc}
    \toprule
    \multirow{2}{*}{\textbf{Team}} &  \multicolumn{3}{c}{\textbf{Reproduction Score (\%)}}  \\
    \cmidrule(lr){2-4}
    & \textbf{Machine Learning} & \textbf{AI4Science} & \textbf{Overall} \\
    \midrule
    \hyphentt{YNU-HPCC-Task11-AgentRep} & 46.69 & 61.25 & 49.64\\
    \hyphentt{zzunlp\_wu} & 22.97 & 31.50 & 24.70 \\
    \underline{\hyphentt{Codex-GPT-5.4}} & 22.89 & 29.26 & 24.19 \\
    \hyphentt{QueenAgent} & 6.80 & 7.89 & 7.03 \\
    \bottomrule
    \end{tabular}
    \end{adjustbox}
    \label{tab:leaderboard}
    \vspace{-2.5em}
\end{table}
\section{Experimental Setup}

\noindent\textbf{Models.}~~For rubric generation, we follow the configuration adopted in prior work~\cite{hong2026llmswritereliablerubrics}. Specifically, we use \hyphentt{Claude Code}~\cite{claude2026claudecode} as the agentic scaffold, \hyphentt{Claude-Sonnet}~\cite{claude2026sonnet4.6card} as the backbone model, and a skill document distilled from \hyphentt{Claude-Opus}~\cite{claude2026opus4.6card}. For evaluation, we use \hyphentt{ChatGPT-4o-mini}~\cite{openai2024gpt4ocard} as the evaluator to reduce computational cost.

\noindent\textbf{Benchmarks.}~~Because \ourworksmall~is released as a training set for participants to develop and refine their models and systems, the main leaderboard is computed on the remaining papers in \ourwork. However, since both human-annotated and model-generated rubrics are available for \ourworksmall, we additionally collect participants' reproduction results on this subset to analyse the alignment between model-generated and human-authored rubrics.

\noindent\textbf{Repositories.}~~We evaluate submissions from three participating teams: \hyphentt{YNU-HPCC-Task11-AgentRep}, \hyphentt{zzunlp\_wu}, and \hyphentt{QueenAgent}. To provide an additional reference point, we also reproduce all 150 papers in \ourwork~using \hyphentt{GPT-5.4}~\cite{singh2026openaigpt5card} with the \hyphentt{Codex} scaffold~\cite{openai2025codex}, which serves as our reproduction baseline. Correlation analyses on \ourworksmall~are conducted using the results from all participating teams and the baseline.

\noindent\textbf{Environment.}~~All experiments are conducted on two NVIDIA L40S GPUs.
\begin{table}[t]
    \centering
    \caption{System performance across rubric types. Each value denotes the percentage of the maximum available score obtained for the corresponding rubric type. \textit{PO}, \textit{PW}, \textit{CI}, \textit{CE}, and \textit{RM} denote \textit{Paper Observation}, \textit{Plan Writing}, \textit{Code Implementation}, \textit{Command Execution}, and \textit{Result Matching}, respectively.}
    \begin{adjustbox}{max width=\textwidth}
    \begin{tabular}{cccccc}
    \toprule
    \multirow{2}{*}{\textbf{Team}} & \multicolumn{5}{c}{\textbf{Obtained Score (\%)}} \\
    \cmidrule(lr){2-6}
    & \textit{PO} & \textit{PW} & \textit{CI} & \textit{CE} & \textit{RM}\\
    \midrule
    \hyphentt{YNU-HPCC-Task11-AgentRep} & 43.73 & 82.45 & 61.89 & 18.04 & 17.24\\
    \hyphentt{zzunlp\_wu} & 46.09 & 35.91 & 29.15 & 10.71 & 5.30 \\
    \underline{\hyphentt{Codex-GPT-5.4}} & 24.07 & 35.12 & 32.37 & 2.84 & 7.67\\
    \hyphentt{QueenAgent} & 52.34 & 3.07 & 0.95 & 0.00 & 0.39 \\
    \bottomrule
    \end{tabular}
    \end{adjustbox}
    \label{tab:obtained_percentage}
\end{table}

\begin{table}[t]
    \vspace{-1.5em}
    \centering
\caption{Correlation between evaluation scores produced by human-annotated and model-generated rubrics on \ourworksmall, with 95\% confidence intervals and summary statistics. \textit{Avg. cnt.} denotes the average number of rubric items per paper, and \textit{Avg. len.} denotes the average criterion length calculated by characters.}
    \begin{adjustbox}{max width=\textwidth}
    \begin{tabular}{ccccc}
    \toprule
    \multirow{2}{*}{\textbf{Rubric Source}} & \multicolumn{2}{c}{\textbf{Correlation Coefficients}} & \multicolumn{2}{c}{\textbf{Statistics}} \\
    \cmidrule(lr){2-3} \cmidrule(lr){4-5} 
    & \textbf{Pearson} & \textbf{Spearman} & \textit{Avg. Cnt.} & \textit{Avg. Len.} \\
    \midrule
    \hyphentt{Human} & - & - & 67.75 & 153.55  \\
    \hyphentt{Model} & 0.93 $\pm$ 0.04 & 0.88 $\pm$ 0.07 & 69.92 & 180.82 \\
    \bottomrule
    \end{tabular}
    \end{adjustbox}
    \label{tab:corr}
    \vspace{-1.5em}
\end{table}
\section{Result Analysis}
\subsection{Leaderboard}

Table~\ref{tab:leaderboard} reports the performance of the participating systems and our reproduction baseline. \hyphentt{YNU-HPCC-Task11-AgentRep} performs best, achieving an overall score of 49.64\%. However, no system exceeds 50\%, and the \hyphentt{Codex-GPT-5.4} baseline reaches only 24.19\%, underscoring the difficulty of reproducing scientific experiments with current LLM-based agents. All systems also perform better on AI4Science papers than on ML papers, possibly because the selected AI4Science papers use more standard ML methods or simpler experimental configurations.

A breakdown by rubric type is provided in Table~\ref{tab:obtained_percentage}, where each percentage represents the score obtained relative to the maximum available for that category. Systems perform relatively well on \textit{Paper Observation}, \textit{Plan Writing}, and \textit{Code Implementation}, but substantially worse on \textit{Command Execution} and \textit{Result Matching}. This gap indicates that execution remains a major bottleneck: current agents can often generate plausible plans and code, yet struggle with runtime failures, environment and dependency issues, and reproducing the reported results. These findings are consistent with prior studies identifying implementation and execution as key limitations of scientific agents~\cite{zhu2025aiscientistsfailstrong,xie2025faraiscientistschanging,hong2026hirashierarchicalmultiagentframework}.

\begin{figure}[t]
    \centering
    \includegraphics[width=\textwidth]{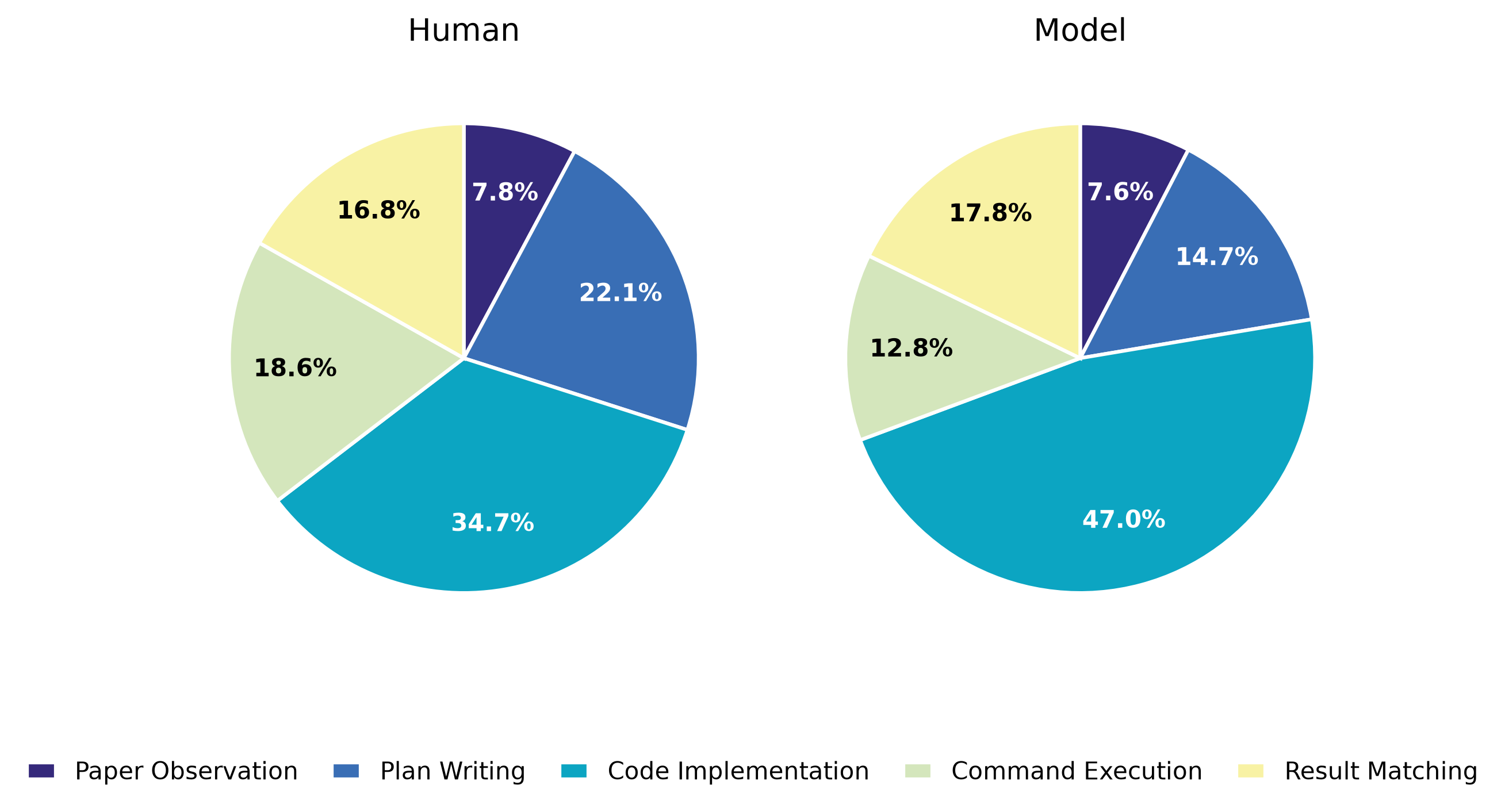}
    \caption{Comparison of score distributions across rubric types for human-annotated and model-generated rubrics.}
    \label{fig:pie_chart}
    \vspace{-2.5em}
\end{figure}

\subsection{Rubric Generation Analysis}
\label{sec:benchmark_analysis}
Table~\ref{tab:corr} reports the correlations between the evaluation scores produced by human-annotated and model-generated rubrics on \ourworksmall, together with their 95\% confidence intervals and summary statistics. The model-generated rubrics achieve Pearson and Spearman correlation coefficients of 0.93 and 0.88, respectively, both with narrow confidence intervals. These results demonstrate strong agreement with the human annotations and support the reliability of the proposed rubric-generation pipeline for scaling the benchmark. The two rubric sources also have similar average numbers of rubric items per paper, suggesting that the generated rubrics broadly capture the structure of the human annotations.

Figure~\ref{fig:pie_chart} further shows that their score distributions are comparable across most rubric types. However, the model-generated criteria are more verbose on average and assign a larger share of the total score to \textit{Code Implementation}, while assigning smaller shares to \textit{Plan Writing} and \textit{Command Execution}. These differences suggest that the generated rubrics could benefit from further calibration of their granularity and category emphasis, consistent with prior findings~\cite{hong2026llmswritereliablerubrics}.

\section{Conclusion}

We introduced \ourwork, a process-oriented benchmark for evaluating agent-based experiment reproduction across ML and AI4Science papers. Our framework combines an MCP-based Action Recorder with fine-grained, paper-specific rubrics, enabling auditable evaluation of the full reproduction process rather than relying solely on final repositories. To support evaluation at scale, we constructed a human-annotated subset and expanded it to 150 papers using an LLM-assisted rubric-generation pipeline across multiple domains. Experimental results show that current agents remain far from reliably reproducing scientific experiments, with execution and result verification emerging as the primary bottlenecks. Meanwhile, the strong agreement between model-generated and human-annotated rubrics supports the scalability of our benchmark construction approach. 

\section*{Limitations}

Although rubric-based evaluation substantially reduces hallucinations, it cannot eliminate them entirely. Due to budget constraints, we use only \hyphentt{GPT-4o-mini} as the evaluator; future work should assess whether stronger models provide more reliable judgments. In addition, our coverage of AI4Science remains preliminary, and extending the benchmark to broader scientific domains is an important direction for future research.
\bibliographystyle{splncs04}
\bibliography{custom}

@misc{starace2025paperbenchevaluatingaisability,
      title={PaperBench: Evaluating AI's Ability to Replicate AI Research}, 
      author={Giulio Starace and Oliver Jaffe and Dane Sherburn and James Aung and Jun Shern Chan and Leon Maksin and Rachel Dias and Evan Mays and Benjamin Kinsella and Wyatt Thompson and Johannes Heidecke and Amelia Glaese and Tejal Patwardhan},
      year={2025},
      eprint={2504.01848},
      archivePrefix={arXiv},
      primaryClass={cs.AI},
      url={https://arxiv.org/abs/2504.01848}, 
}

@misc{seo2025paper2codeautomatingcodegeneration,
      title={Paper2Code: Automating Code Generation from Scientific Papers in Machine Learning}, 
      author={Minju Seo and Jinheon Baek and Seongyun Lee and Sung Ju Hwang},
      year={2025},
      eprint={2504.17192},
      archivePrefix={arXiv},
      primaryClass={cs.CL},
      url={https://arxiv.org/abs/2504.17192}, 
}

@misc{zhao2025autoreproduceautomaticaiexperiment,
      title={AutoReproduce: Automatic AI Experiment Reproduction with Paper Lineage}, 
      author={Xuanle Zhao and Zilin Sang and Yuxuan Li and Qi Shi and Weilun Zhao and Shuo Wang and Duzhen Zhang and Xu Han and Zhiyuan Liu and Maosong Sun},
      year={2025},
      eprint={2505.20662},
      archivePrefix={arXiv},
      primaryClass={cs.AI},
      url={https://arxiv.org/abs/2505.20662}, 
}

@misc{openai2024gpt4ocard,
      title={GPT-4o System Card}, 
      author={OpenAI},
      year={2024},
      eprint={2410.21276},
      archivePrefix={arXiv},
      primaryClass={cs.CL},
      url={https://arxiv.org/abs/2410.21276}, 
}

@misc{openai2025codex,
      title={Introducing Codex}, 
      author={OpenAI},
      year={2025},
      url={https://openai.com/index/introducing-codex/}, 
}

@misc{claude2026sonnet4.6card,
      title={Claude-Sonnet System Card}, 
      author={Anthropic},
      year={2026},
      url={http://anthropic.com/claude-sonnet-4-6-system-card}, 
}

@misc{claude2026opus4.6card,
      title={Claude-Opus System Card}, 
      author={Anthropic},
      year={2026},
      url={https://www-cdn.anthropic.com/6a5fa276ac68b9aeb0c8b6af5fa36326e0e166dd.pdf}, 
}

@misc{claude2026claudecode,
      title={Claude Code Overview}, 
      author={Anthropic},
      year={2026},
      url={
https://code.claude.com/docs}, 
}

@inproceedings{magnusson-etal-2023-reproducibility,
    title = "Reproducibility in {NLP}: What Have We Learned from the Checklist?",
    author = "Magnusson, Ian  and
      Smith, Noah A.  and
      Dodge, Jesse",
    editor = "Rogers, Anna  and
      Boyd-Graber, Jordan  and
      Okazaki, Naoaki",
    booktitle = "Findings of the Association for Computational Linguistics: ACL 2023",
    month = jul,
    year = "2023",
    address = "Toronto, Canada",
    publisher = "Association for Computational Linguistics",
    url = "https://aclanthology.org/2023.findings-acl.809/",
    doi = "10.18653/v1/2023.findings-acl.809",
    pages = "12789--12811"
}

@misc{xie2025faraiscientistschanging,
      title={How Far Are AI Scientists from Changing the World?}, 
      author={Qiujie Xie and Yixuan Weng and Minjun Zhu and Fuchen Shen and Shulin Huang and Zhen Lin and Jiahui Zhou and Zilan Mao and Zijie Yang and Linyi Yang and Jian Wu and Yue Zhang},
      year={2025},
      eprint={2507.23276},
      archivePrefix={arXiv},
      primaryClass={cs.AI},
      url={https://arxiv.org/abs/2507.23276}, 
}

@article{DBLP:journals/corr/abs-2408-06292,
  publtype={informal},
  author={Chris Lu and Cong Lu and Robert Tjarko Lange and Jakob N. Foerster and Jeff Clune and David Ha},
  title={The AI Scientist: Towards Fully Automated Open-Ended Scientific Discovery},
  year={2024},
  cdate={1704067200000},
  journal={CoRR},
  volume={abs/2408.06292},
  url={https://doi.org/10.48550/arXiv.2408.06292}
}

@misc{zhu2025aiscientistsfailstrong,
      title={AI Scientists Fail Without Strong Implementation Capability}, 
      author={Minjun Zhu and Qiujie Xie and Yixuan Weng and Jian Wu and Zhen Lin and Linyi Yang and Yue Zhang},
      year={2025},
      eprint={2506.01372},
      archivePrefix={arXiv},
      primaryClass={cs.AI},
      url={https://arxiv.org/abs/2506.01372}, 
}

@article{doi:10.1142/S0218194022500358,
author = {Lin, Jialiang and Wang, Yingmin and Yu, Yao and Zhou, Yu and Chen, Yidong and Shi, Xiaodong},
title = {Automatic Analysis of Available Source Code of Top Artificial Intelligence Conference Papers},
journal = {International Journal of Software Engineering and Knowledge Engineering},
volume = {32},
number = {07},
pages = {947-970},
year = {2022},
doi = {10.1142/S0218194022500358},
URL = { https://doi.org/10.1142/S0218194022500358},
eprint = { https://doi.org/10.1142/S0218194022500358}
}

@inproceedings{yuan-etal-2025-dolphin,
    title = "Dolphin: Moving Towards Closed-loop Auto-research through Thinking, Practice, and Feedback",
    author = "Yuan, Jiakang  and
      Yan, Xiangchao  and
      Zhang, Bo  and
      Chen, Tao  and
      Shi, Botian  and
      Ouyang, Wanli  and
      Qiao, Yu  and
      Bai, Lei  and
      Zhou, Bowen",
    editor = "Che, Wanxiang  and
      Nabende, Joyce  and
      Shutova, Ekaterina  and
      Pilehvar, Mohammad Taher",
    booktitle = "Proceedings of the 63rd Annual Meeting of the Association for Computational Linguistics (Volume 1: Long Papers)",
    month = jul,
    year = "2025",
    address = "Vienna, Austria",
    publisher = "Association for Computational Linguistics",
    url = "https://aclanthology.org/2025.acl-long.1056/",
    doi = "10.18653/v1/2025.acl-long.1056",
    pages = "21768--21789",
    ISBN = "979-8-89176-251-0"
}

@misc{hong2025onesizefitsallinversionlearninghighly,
      title={Beyond One-Size-Fits-All: Inversion Learning for Highly Effective NLG Evaluation Prompts}, 
      author={Hanhua Hong and Chenghao Xiao and Yang Wang and Yiqi Liu and Wenge Rong and Chenghua Lin},
      year={2025},
      eprint={2504.21117},
      archivePrefix={arXiv},
      primaryClass={cs.CL},
      url={https://arxiv.org/abs/2504.21117}, 
}

@misc{hong2026hirashierarchicalmultiagentframework,
      title={HiRAS: A Hierarchical Multi-Agent Framework for Paper-to-Code Generation and Execution}, 
      author={Hanhua Hong and Yizhi LI and Jiaoyan Chen and Sophia Ananiadou and Xiaoli Li and Jung-jae Kim and Chenghua Lin},
      year={2026},
      eprint={2604.17745},
      archivePrefix={arXiv},
      primaryClass={cs.CL},
      url={https://arxiv.org/abs/2604.17745}, 
}

@misc{lyu2026evoscientistmultiagentevolvingai,
      title={EvoScientist: Towards Multi-Agent Evolving AI Scientists for End-to-End Scientific Discovery}, 
      author={Yougang Lyu and Xi Zhang and Xinhao Yi and Yuyue Zhao and Shuyu Guo and Wenxiang Hu and Jan Piotrowski and Jakub Kaliski and Jacopo Urbani and Zaiqiao Meng and Lun Zhou and Xiaohui Yan},
      year={2026},
      eprint={2603.08127},
      archivePrefix={arXiv},
      primaryClass={cs.CL},
      url={https://arxiv.org/abs/2603.08127}, 
}

@misc{arora2025healthbenchevaluatinglargelanguage,
      title={HealthBench: Evaluating Large Language Models Towards Improved Human Health}, 
      author={Rahul K. Arora and Jason Wei and Rebecca Soskin Hicks and Preston Bowman and Joaquin Quiñonero-Candela and Foivos Tsimpourlas and Michael Sharman and Meghan Shah and Andrea Vallone and Alex Beutel and Johannes Heidecke and Karan Singhal},
      year={2025},
      eprint={2505.08775},
      archivePrefix={arXiv},
      primaryClass={cs.CL},
      url={https://arxiv.org/abs/2505.08775}, 
}

@misc{zhang2026rubricbenchaligningmodelgeneratedrubrics,
      title={RubricBench: Aligning Model-Generated Rubrics with Human Standards}, 
      author={Qiyuan Zhang and Junyi Zhou and Yufei Wang and Fuyuan Lyu and Yidong Ming and Can Xu and Qingfeng Sun and Kai Zheng and Peng Kang and Xue Liu and Chen Ma},
      year={2026},
      eprint={2603.01562},
      archivePrefix={arXiv},
      primaryClass={cs.AI},
      url={https://arxiv.org/abs/2603.01562}, 
}

@inproceedings{
gunjal2026rubrics,
title={Rubrics as Rewards: Reinforcement Learning Beyond Verifiable Domains},
author={Anisha Gunjal and Anthony Wang and Elaine Lau and Vaskar Nath and Yunzhong He and Bing Liu and Sean M. Hendryx},
booktitle={The Fourteenth International Conference on Learning Representations},
year={2026},
url={https://openreview.net/forum?id=c1bTcrDmt4}
}

@misc{liu2026openrubricsscalablesyntheticrubric,
      title={OpenRubrics: Towards Scalable Synthetic Rubric Generation for Reward Modeling and LLM Alignment}, 
      author={Tianci Liu and Ran Xu and Tony Yu and Ilgee Hong and Carl Yang and Tuo Zhao and Haoyu Wang},
      year={2026},
      eprint={2510.07743},
      archivePrefix={arXiv},
      primaryClass={cs.CL},
      url={https://arxiv.org/abs/2510.07743}, 
}

@misc{singh2026openaigpt5card,
      title={OpenAI GPT-5 System Card}, 
      author={Aaditya Singh and Adam Fry and Adam Perelman and Adam Tart and Adi Ganesh and others},
      year={2026},
      eprint={2601.03267},
      archivePrefix={arXiv},
      primaryClass={cs.CL},
      url={https://arxiv.org/abs/2601.03267}, 
}

@article{abramson_accurate_2024,
	title = {Accurate structure prediction of biomolecular interactions with {AlphaFold} 3},
	volume = {630},
	issn = {1476-4687},
	url = {https://doi.org/10.1038/s41586-024-07487-w},
	doi = {10.1038/s41586-024-07487-w},
	number = {8016},
	journal = {Nature},
	author = {Abramson, Josh and Adler, Jonas and Dunger, Jack and Evans, Richard and Green, Tim and Pritzel, Alexander and Ronneberger, Olaf and Willmore, Lindsay and Ballard, Andrew J. and Bambrick, Joshua and Bodenstein, Sebastian W. and Evans, David A. and Hung, Chia-Chun and O’Neill, Michael and Reiman, David and Tunyasuvunakool, Kathryn and Wu, Zachary and Žemgulytė, Akvilė and Arvaniti, Eirini and Beattie, Charles and Bertolli, Ottavia and Bridgland, Alex and Cherepanov, Alexey and Congreve, Miles and Cowen-Rivers, Alexander I. and Cowie, Andrew and Figurnov, Michael and Fuchs, Fabian B. and Gladman, Hannah and Jain, Rishub and Khan, Yousuf A. and Low, Caroline M. R. and Perlin, Kuba and Potapenko, Anna and Savy, Pascal and Singh, Sukhdeep and Stecula, Adrian and Thillaisundaram, Ashok and Tong, Catherine and Yakneen, Sergei and Zhong, Ellen D. and Zielinski, Michal and Žídek, Augustin and Bapst, Victor and Kohli, Pushmeet and Jaderberg, Max and Hassabis, Demis and Jumper, John M.},
	month = jun,
	year = {2024},
	pages = {493--500},
}

@misc{bran2023chemcrowaugmentinglargelanguagemodels,
      title={ChemCrow: Augmenting large-language models with chemistry tools}, 
      author={Andres M Bran and Sam Cox and Oliver Schilter and Carlo Baldassari and Andrew D White and Philippe Schwaller},
      year={2023},
      eprint={2304.05376},
      archivePrefix={arXiv},
      primaryClass={physics.chem-ph},
      url={https://arxiv.org/abs/2304.05376}, 
}

@misc{baker20161,
  title={1,500 scientists lift the lid on reproducibility},
  author={Baker, Monya},
  year={2016},
  publisher={Nature Publishing Group UK London}
}

@article{siegel2024corebench,
  title={{CORE-Bench}: Fostering the Credibility of Published Research Through a Computational Reproducibility Agent Benchmark},
  author={Siegel, Zachary S. and Kapoor, Sayash and Nadgir, Nitya and Stroebl, Benedikt and Narayanan, Arvind},
  journal={Transactions on Machine Learning Research},
  year={2025},
  month={01},
  url={https://openreview.net/forum?id=BsMMc4MEGS}
}

@inproceedings{autoexperiment,
  title={From Reproduction to Replication: Evaluating Research Agents with Progressive Code Masking},
  author={Kim, Gyeongwon James and Wilf, Alex and Morency, Louis-Philippe and Fried, Daniel},
  booktitle={Proceedings of the International Conference on Learning Representations (ICLR)},
  year={2026},
  address={Pittsburgh, PA, USA},
  organization={Carnegie Mellon University}
}

@inproceedings{xiang2025scireplicate,
  title={{SciReplicate-Bench}: Benchmarking LLMs in Agent-driven Algorithmic Reproduction from Research Papers},
  author={Xiang, Yanzheng and Yan, Hanqi and Ouyang, Shuyin and Gui, Lin and He, Yulan},
  booktitle={Proceedings of the Conference on Language Modeling (COLM)},
  year={2025},
  organization={King's College London and The Alan Turing Institute}
}

@inproceedings{yan-etal-2025-lmr,
    title = "{LMR}-{BENCH}: Evaluating {LLM} Agent{'}s Ability on Reproducing Language Modeling Research",
    author = "Yan, Shuo  and
      Li, Ruochen  and
      Luo, Ziming  and
      Wang, Zimu  and
      Li, Daoyang  and
      Jing, Liqiang  and
      He, Kaiyu  and
      Wu, Peilin  and
      Ni, Juntong  and
      Michalopoulos, George  and
      Zhang, Yue  and
      Zhang, Ziyang  and
      Zhang, Mian  and
      Chen, Zhiyu  and
      Du, Xinya",
    editor = "Christodoulopoulos, Christos  and
      Chakraborty, Tanmoy  and
      Rose, Carolyn  and
      Peng, Violet",
    booktitle = "Proceedings of the 2025 Conference on Empirical Methods in Natural Language Processing",
    month = nov,
    year = "2025",
    address = "Suzhou, China",
    publisher = "Association for Computational Linguistics",
    url = "https://aclanthology.org/2025.emnlp-main.314/",
    doi = "10.18653/v1/2025.emnlp-main.314",
    pages = "6164--6186",
    ISBN = "979-8-89176-332-6",
}

@inproceedings{chan2024mlebench,
  title={{MLE-bench}: Evaluating Machine Learning Agents on Machine Learning Engineering},
  author={Chan, Jun Shern and Chowdhury, Neil and Jaffe, Oliver and Aung, James and Sherburn, Dane and Mays, Evan and Starace, Giulio and Liu, Kevin and Maksin, Leon and Patwardhan, Tejal and Weng, Lilian and M{\k{a}}dry, Aleksander},
  booktitle={Proceedings of the International Conference on Learning Representations (ICLR)},
  year={2025}
}

@inproceedings{huang2024mlagentbench,
  title={{MLAgentBench}: Evaluating Language Agents on Machine Learning Experimentation},
  author={Huang, Qian and Vora, Jian and Liang, Percy and Leskovec, Jure},
  booktitle={Proceedings of the 41st International Conference on Machine Learning (ICML)},
  year={2024},
  pages={20271--20309},
  articleno={814}
}

@article{rubrichub2026,
  title={RubricHub: A Comprehensive and Highly Discriminative Rubric Dataset via Automated Coarse-to-Fine Generation},
  author={},
  journal={arXiv preprint arXiv:2601.08430},
  year={2026}
}

@inproceedings{rubricisallyouneed2025,
  title={Rubric Is All You Need: Improving LLM-Based Code Evaluation With Question-Specific Rubrics},
  author={},
  booktitle={Proceedings of the 2025 ACM Conference on International Computing Education Research V.1},
  year={2025},
  doi={10.1145/3702652.3744220}
}

@inproceedings{wang2023selfinstruct,
  title={Self-Instruct: Aligning Language Models with Self-Generated Instructions},
  author={Wang, Yizhong and Kordi, Yeganeh and Mishra, Swaroop and Liu, Alisa and Smith, Noah A. and Khashabi, Daniel and Hajishirzi, Hannaneh},
  booktitle={Proceedings of the 61st Annual Meeting of the Association for Computational Linguistics (Volume 1: Long Papers)},
  year={2023}
}

@article{xu2023wizardlm,
  title={{WizardLM}: Empowering Large Language Models to Follow Complex Instructions},
  author={Xu, Can and Sun, Qingfeng and Zheng, Kai and Geng, Xiubo and Zhao, Pu and Feng, Jiazhan and Tao, Chongyang and Jiang, Daxin},
  journal={arXiv preprint arXiv:2304.12244},
  year={2023}
}

@article{gao2023gllava,
  title={{G-LLaVA}: Solving Geometric Problem with Multi-Modal Large Language Model},
  author={Gao, Jiahui and Pi, Renjie and Zhang, Jipeng and Ye, Jiacheng and Zhong, Wanjun and Wang, Yufei and Hong, Lanqing and Han, Jianhua and Xu, Hang and Li, Zhenguo and Kong, Lingpeng},
  journal={arXiv preprint arXiv:2312.11370},
  year={2023}
}

@article{zhou2025textda,
  title={Text Data Augmentation for Large Language Models: A Comprehensive Survey of Methods, Challenges, and Opportunities},
  author={},
  journal={Artificial Intelligence Review},
  year={2025}
}

@article{wang2026mineru2,
  title={MinerU2. 5-Pro: Pushing the Limits of Data-Centric Document Parsing at Scale},
  author={Wang, Bin and He, Tianyao and Ouyang, Linke and Wu, Fan and Zhao, Zhiyuan and Chu, Tao and Qu, Yuan and Jin, Zhenjiang and Zeng, Weijun and Miao, Ziyang and others},
  journal={arXiv preprint arXiv:2604.04771},
  year={2026}
}

@misc{hong2026llmswritereliablerubrics,
      title={Can LLMs Write Reliable Rubrics? A Meta-Evaluation for Experiment Reproduction}, 
      author={Hanhua Hong and Yizhi Li and Jiaoyan Chen and Luu Gia Huy and Sophia Ananiadou and Jung-jae Kim and Chenghua Lin},
      year={2026},
      eprint={2607.12835},
      archivePrefix={arXiv},
      primaryClass={cs.CL},
      url={https://arxiv.org/abs/2607.12835}, 
}

\clearpage
\appendix
\section{Log Example}
\label{app:log_example}
\begin{figure}[b]
\centering
\tcbset{
    colback=gray!5!white,
    width=0.95\textwidth,
    fontupper=\small,
    left=1pt,
    right=1pt,
    valign=center,
    before=\vspace{0pt},
    after=\vspace{0pt}
}
\begin{tcolorbox}[]
[\\
\tcbtab\{\\
\tcbtab\tcbtab"id": 1,\\
\tcbtab\tcbtab"timestamp": ...,\\
\tcbtab\tcbtab"tool": "Read",\\
\tcbtab\tcbtab"arguments": \{\\
\tcbtab\tcbtab\tcbtab"path": "paper.md",\\
\tcbtab\tcbtab\tcbtab"start\_offset": null,\\
\tcbtab\tcbtab\tcbtab"end\_offset": null\\
\tcbtab\tcbtab\},\\
\tcbtab\tcbtab"result": \{\\
\tcbtab\tcbtab\tcbtab"content": "\# My Paper\textbackslash n\#\# Abstract \textbackslash n...",\\
\tcbtab\tcbtab\},\\
\tcbtab\{\\
\tcbtab\tcbtab"id": 2,\\
\tcbtab\tcbtab"timestamp": "...",\\
\tcbtab\tcbtab"tool": "Write",\\
\tcbtab\tcbtab"arguments": \{\\
\tcbtab\tcbtab\tcbtab"path": "model.py",\\
\tcbtab\tcbtab\tcbtab"content": "import torch\textbackslash nimport torch.nn as nn\textbackslash n..."\\
\tcbtab\tcbtab\},\\
\tcbtab\tcbtab\tcbtab"result": \{\\
\tcbtab\tcbtab\tcbtab"success": true\\
\tcbtab\tcbtab\},\\
\tcbtab,\\
\tcbtab\{\\
\tcbtab\tcbtab"id": 3,\\
\tcbtab\tcbtab"timestamp": "...",\\
\tcbtab\tcbtab"tool": "Execute",\\
\tcbtab\tcbtab"arguments": \{\\
\tcbtab\tcbtab\tcbtab"cmd": "python train.py --epochs 10"\\
\tcbtab\tcbtab\},
\tcbtab\tcbtab"result": \{\\
\tcbtab\tcbtab\tcbtab"stdout": "Epoch 1/10 loss=0.45\textbackslash nEpoch 2/10 loss=0.32\textbackslash n...",\\
\tcbtab\tcbtab\tcbtab"stderr": "",\\
\tcbtab\tcbtab\tcbtab"success": true\\
\tcbtab\tcbtab\},\\
\tcbtab\}\\
]
\end{tcolorbox}
\vspace{-0.5em}
\caption{An example of Action Recorder logs.}
\label{fig:json_example}
\end{figure}

\end{document}